\documentclass[conference]{IEEEtran}
\IEEEoverridecommandlockouts
\usepackage{cite}
\usepackage{amsmath,amssymb,amsfonts}
\usepackage{algorithmic}
\usepackage{graphicx}
\usepackage{textcomp}
\usepackage{xcolor}
\usepackage{booktabs}   
\usepackage{multirow}   
\usepackage{array}      
\usepackage{tabularx}
\usepackage{makecell}
\usepackage{enumitem}
\usepackage[
  colorlinks=true,
  linkcolor=black,
  citecolor=black,
  urlcolor=blue
]{hyperref}
\def\BibTeX{{\rm B\kern-.05em{\sc i\kern-.025em b}\kern-.08em
    T\kern-.1667em\lower.7ex\hbox{E}\kern-.125emX}}
\begin{document}

\newcommand{\model}{\textsc{FAIRPLAI}}

\title{\model: A Human-in-the-Loop Approach to Fair and Private Machine Learning}

\author{
\IEEEauthorblockN{
David Sanchez\IEEEauthorrefmark{1}$^*$, 
Holly Lopez\IEEEauthorrefmark{2}$^*$, 
Michelle Buraczyk\IEEEauthorrefmark{3}$^*$,
Anantaa Kotal\IEEEauthorrefmark{1}, 
}
\IEEEauthorblockA{
\IEEEauthorrefmark{1} Dept. of Computer Science, The University of Texas at El Paso, dasanchez13@miners.utep.edu, akotal@utep.edu\\
\IEEEauthorrefmark{2} Dept. of Mathematics, Mountain View High School, holly.lopez@clint.net\\
\IEEEauthorrefmark{3} Dept. of Mathematics, El Paso Independent School District, myburacz@episd.org
}
}

\maketitle
\def\thefootnote{*}\footnotetext{These authors contributed equally to this work}\def\thefootnote{\arabic{footnote}}

\begin{abstract}
As machine learning systems move from theory to practice, they are increasingly tasked with decisions that affect healthcare access, financial opportunities, hiring, and public services. In these contexts, accuracy is only one piece of the puzzle---models must also be fair to different groups, protect individual privacy, and remain accountable to stakeholders. Achieving all three is difficult: differential privacy can unintentionally worsen disparities, fairness interventions often rely on sensitive data that privacy restricts, and automated pipelines ignore that fairness is ultimately a \textbf{human and contextual judgment}. We introduce \textit{\model: (Fair and Private Learning with Active Human Influence)}, a practical framework that integrates human oversight into the design and deployment of machine learning systems. \model~ works in three ways: (1) it constructs privacy--fairness frontiers that make trade-offs between accuracy, privacy guarantees, and group outcomes transparent; (2) it enables interactive stakeholder input, allowing decision-makers to select fairness criteria and operating points that reflect their domain needs; and (3) it embeds a differentially private auditing loop, giving humans the ability to review explanations and edge cases without compromising individual data security.  Applied to benchmark datasets, \model~ consistently preserves strong privacy protections while reducing fairness disparities relative to automated baselines. More importantly, it provides a straightforward, interpretable process for practitioners to manage competing demands of accuracy, privacy, and fairness in socially impactful applications. By embedding human judgment where it matters most, \model~ offers a pathway to machine learning systems that are effective, responsible, and trustworthy in practice. Github: \href{github.com/Li1Davey/Fairplai}{\url{https://github.com/Li1Davey/Fairplai}}
\end{abstract}

\begin{IEEEkeywords}
Human-in-the-loop AI, Trustworthy AI, Fair Algorithms, Differential Privacy

\end{IEEEkeywords}

\section{Introduction}

Machine learning has moved from theoretical development to widespread practical deployment in domains directly impacting people’s lives. Algorithms now assist in determining eligibility for loans, allocating healthcare resources, screening job candidates, and guiding public safety policies. In high-stakes applications, accuracy---while essential---is no longer sufficient. Increasingly, there is recognition that machine learning systems must also be \textbf{fair}, ensuring equitable treatment across groups; \textbf{private}, protecting sensitive individual data; and \textbf{accountable}, enabling trust and oversight. Yet growing evidence indicates that today’s systems often fall short of these standards.  

Recent research has demonstrated that machine learning systems can reproduce and even amplify existing social inequities when deployed in practice. For example, a large-scale evaluation of a widely used healthcare risk prediction system revealed systematic underestimation of the health needs of Black patients compared to white patients with equivalent medical conditions \cite{obermeyer2019dissecting}. In employment settings, automated hiring platforms have been found to reflect gendered patterns in candidate selection. At the same time, facial recognition systems have shown disproportionately high error rates for women and individuals with darker skin tones \cite{buolamwini2018gender}. These findings underscore how seemingly neutral models can generate disparate outcomes across demographic groups.  

At the same time, mounting evidence highlights the vulnerability of machine learning models to privacy attacks. Studies have shown that adversaries can determine whether an individual’s record was part of a model’s training set through \textit{membership inference attacks}, or reconstruct sensitive personal attributes through \textit{model inversion}. Such vulnerabilities expose individuals to risks that extend beyond technical failures, undermining the system's reliability and the legitimacy of its deployment in sensitive domains.  

These challenges crystallize into what we term the \textbf{fairness--privacy--accuracy paradox}. Differential privacy provides strong protection against data leakage by injecting carefully calibrated randomness during training. However, this same noise can disproportionately harm underrepresented groups with weak signals, thereby increasing disparities. Conversely, fairness-enhancing methods often rely on access to sensitive demographic attributes such as race, gender, or age---information that privacy-preserving frameworks seek to limit or conceal. Meanwhile, optimizing purely for predictive accuracy without consideration of fairness or privacy can exacerbate both inequities and exposure risks. Striving for one goal often undermines another, and balancing all three has proven deeply challenging.  

Much of the existing research on fairness and privacy attempts to navigate this paradox through algorithmic innovation---developing new objective functions, constraints, or optimization procedures. While these methods offer valuable theoretical insights, they frequently share a key limitation: they treat fairness and privacy as \textbf{abstract optimization targets}. In practice, however, fairness is not a single universal definition, but a collection of normative judgments that vary across contexts and stakeholders. Similarly, privacy is not solely a mathematical guarantee, but also a social expectation shaped by human values and institutional trust. Treating these concerns as purely technical quantities risks missing the fact that decisions about what is ``fair enough'' or ``private enough'' are fundamentally human judgments.  

This observation motivates our central \textbf{research question}: \textit{How can machine learning systems balance fairness, privacy, and accuracy in ways that reflect human notions of morality, security, and accountability, rather than abstract optimization alone?}  

To address this question, we propose \model\ (\textit{Fair and Private Learning with Active Human Influence}), a \textbf{human-in-the-loop framework} for fair and private machine learning. Instead of seeking a single automated solution to reconcile fairness and privacy, \model\ explicitly incorporates human oversight at key points in the machine learning lifecycle. It does so through three main components:  

\begin{enumerate}
    \item \textbf{Privacy--fairness frontiers}: \model\ generates interpretable visualizations that map the trade-offs between predictive accuracy, fairness disparities, and privacy budgets, enabling stakeholders to see how different design choices affect outcomes.  
    \item \textbf{Interactive stakeholder selection}: Rather than fixing one fairness definition in advance, \model\ allows domain experts, policymakers, or organizational leaders to select criteria (e.g., demographic parity, equalized odds, or counterfactual fairness) and operating points that best reflect their context.    
\end{enumerate}  

Through these mechanisms, \model\ reframes fairness and privacy not as constraints to be optimized away, but as values to be \textbf{contextualized, negotiated, and overseen by humans}. The goal is not a purely theoretical optimum, but a practical balance that aligns technical safeguards with how humans reason about fairness and security.  

The contributions of this paper are threefold:  

\begin{itemize}
    \item We articulate the fairness--privacy--accuracy paradox, showing why existing automated methods struggle to reconcile these objectives in real-world settings.  
    \item We introduce \model, a human-in-the-loop framework that integrates differential privacy with fairness-aware modeling while making trade-offs explicit through privacy--fairness frontiers.  
    \item We empirically demonstrate that \model\ preserves rigorous privacy guarantees, reduces fairness disparities compared to automated baselines, and provides interpretable decision points for human oversight in deployment.  
\end{itemize}  

By embedding human judgment directly into the learning process, \model\ advances machine learning systems that are not only accurate, but also \textbf{fair, private, and accountable to the values of those they serve}.

\section{Problem Statement: The Fairness–Privacy–Accuracy Paradox}
Machine learning systems deployed in socially sensitive domains are judged not only by how well they predict, but also by how responsibly they handle data and how equitably they treat individuals and groups. This means that three objectives must be considered together:

\subsection{Performance (Accuracy)}

Performance, often expressed as predictive accuracy, is the most established measure of machine learning. In supervised learning, models are trained on datasets consisting of input features and corresponding outcomes, and their quality is assessed by how well predictions align with the true outcomes.  

The simplest performance measure is accuracy, defined as the proportion of correct predictions:
\[
\text{Accuracy} = \frac{\text{Number of correct predictions}}{\text{Total predictions}}.
\]  

Depending on the task, other performance metrics may be more informative. In medical applications, \textit{recall} (the fraction of actual positive cases correctly identified) is often prioritized, since missing a diagnosis can have serious consequences. In domains like fraud detection or spam filtering, \textit{precision} (the fraction of positive predictions that are truly correct) may be more critical, to avoid overwhelming human reviewers with false alarms. For imbalanced datasets, composite measures such as the \textit{F1 score} or the \textit{area under the ROC curve (AUC)} are widely used.

\subsection{Privacy}

Privacy in machine learning refers to protecting sensitive information about individuals whose data is used to train models. Standard training methods can unintentionally reveal details about training data, for example, through \textit{membership inference attacks}, which test whether a particular record was included in training, or \textit{model inversion attacks}, which attempt to reconstruct sensitive features from model outputs. These risks highlight the need for formal privacy guarantees.  

\textbf{Differential privacy (DP)} is the most widely adopted standard for such guarantees. Intuitively, it ensures that the output of an algorithm does not change significantly when a single individual’s data is added or removed, thereby limiting what can be inferred about that individual.  Formally, a randomized algorithm $\mathcal{M}$ satisfies $(\varepsilon, \delta)$-differential privacy if, for any two datasets $D$ and $D'$ that differ in exactly one record, and for any possible output set $S$:  
\[
\Pr[\mathcal{M}(D) \in S] \leq e^\varepsilon \Pr[\mathcal{M}(D') \in S] + \delta.
\]  

Here, $\varepsilon$ is called the \textit{privacy budget} and quantifies the maximum outcome difference between neighboring datasets. Smaller values of $\varepsilon$ indicate stronger privacy guarantees, while $\delta$ accounts for a small probability that the guarantee does not hold. In practice, differential privacy is implemented by introducing controlled randomness, such as adding noise to statistics or gradients during model training. This randomness masks the contribution of any single individual, ensuring that their presence in the data cannot be confidently determined.  

\subsection{Fairness}

Fairness in machine learning concerns the equitable treatment of individuals and groups in the predictions and decisions generated by models. Without explicit safeguards, algorithms trained on historical data often reproduce existing social biases, leading to systematic disparities in outcomes. For example, a loan approval system might disproportionately reject applications from certain demographic groups, or a hiring algorithm might rank candidates differently on the basis of gender.  

To study and mitigate such disparities, researchers have introduced several formal definitions of fairness. While the terminology varies, most definitions fall under the category of \textit{group fairness}, which evaluates whether predictions differ across groups defined by a sensitive attribute $A$ (such as gender, race, or age). Some of the most widely used criteria include:  

\begin{itemize}
    \item \textbf{Disparate Impact Ratio (DIR):}  
    This criterion evaluates the ratio of positive prediction rates between protected and reference groups. A common rule of thumb, known as the ``80\% rule,'' considers a classifier fair if the protected group receives positive outcomes at least 80\% as often as the reference group:  
    \[
    \frac{P(\hat{Y} = 1 \mid A = 1)}{P(\hat{Y} = 1 \mid A = 0)} \geq 0.8.
    \] 
    
    \item \textbf{Demographic Parity (DP):}  
    A classifier satisfies demographic parity if the probability of receiving a positive prediction is the same for all groups. Formally,  
    \[
    P(\hat{Y} = 1 \mid A = 0) = P(\hat{Y} = 1 \mid A = 1).
    \]  
    This criterion ensures that outcomes are independent of group membership.  

    \item \textbf{Equalized Odds (EO):}  
    A stronger condition, equalized odds, requires that prediction errors are distributed equally across groups. Specifically, both the true positive rate and false positive rate must be the same for each group:  
    \begin{align*}
        P(\hat{Y} = 1 \mid Y = y, A = 0) = \\
        P(\hat{Y} = 1 \mid Y = y, A = 1), \quad y \in \{0,1\}.
    \end{align*} 

    \item \textbf{Equal Opportunity (EOpp):}  
    A relaxation of equalized odds, equal opportunity requires only that the true positive rate is equal across groups:  
    \[
    P(\hat{Y} = 1 \mid Y = 1, A = 0) = P(\hat{Y} = 1 \mid Y = 1, A = 1).
    \]
\end{itemize}  

These definitions highlight that fairness is not a single concept but a family of criteria, each emphasizing different aspects of equity. Importantly, many of these criteria are mutually incompatible; it is often impossible to satisfy all of them simultaneously in a single model. As a result, the choice of which fairness definition to enforce depends on the domain and the values of the stakeholders involved.  

\begin{figure}[ht]
    \centering
    \includegraphics[width=\columnwidth]
    {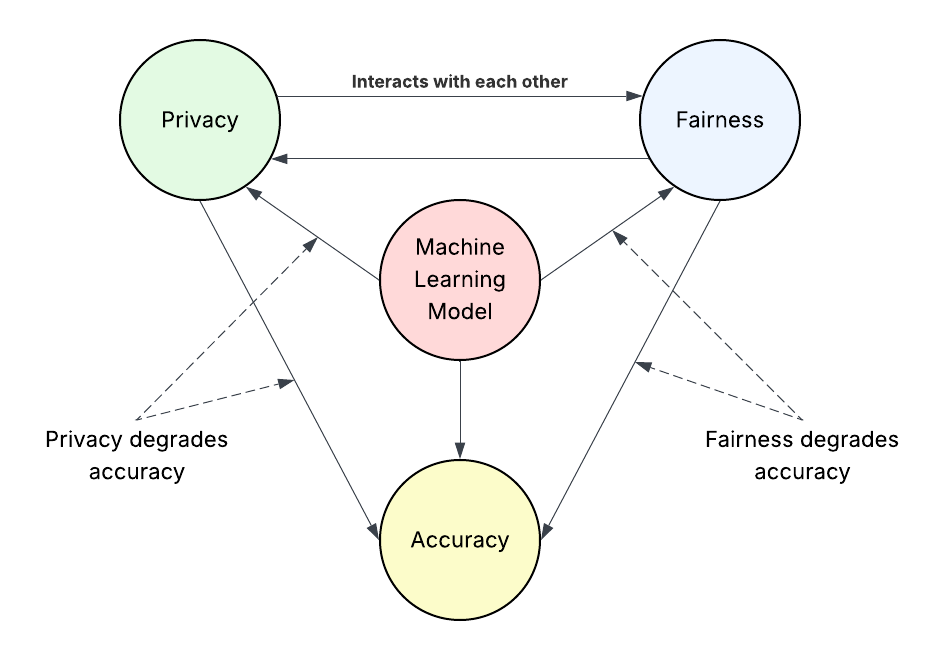}
    \caption{Fairness--Privacy--Accuracy Paradox in Machine Learning Models}
    \label{fig:reident}
\end{figure}

\subsection{The Fairness--Privacy--Accuracy Paradox}

The three dimensions we have described --- performance, privacy, and fairness --- are each essential for responsible machine learning. However, when considered together, they form a paradox: improving one often undermines another. This tension is not merely theoretical but has been observed across domains where machine learning systems are deployed.  

\textbf{Performance vs. Privacy.}  
To guarantee privacy, methods such as differential privacy introduce randomness (e.g., adding noise to training or outputs) so that no individual’s data can be distinguished. While this protects confidentiality, the added noise inevitably reduces predictive precision. For instance, a disease risk prediction model trained under strict privacy guarantees may be less accurate in distinguishing between high- and low-risk patients. In practice, this means fewer correct diagnoses, even though patient privacy is preserved.  

\textbf{Performance vs. Fairness.}  
A model optimized only for accuracy may inadvertently favor the majority group represented in the data, while producing higher error rates for underrepresented groups. For example, a hiring model trained to maximize predictive performance on historical applications may select candidates with high overall accuracy but disproportionately misclassify qualified women or minority applicants. Here, pursuing maximum accuracy leads to inequitable outcomes.  

\textbf{Fairness vs. Privacy.}  
Many fairness interventions rely on access to sensitive attributes such as race, gender, or age to detect and correct disparities. Yet privacy mechanisms are designed to minimize the use or disclosure of these attributes. Consider a credit-scoring model: ensuring that loan approval rates are equal across groups requires explicit knowledge of applicants’ demographic information. However, a strict privacy regime may restrict the use of this information, making fairness adjustments more difficult or impossible.  

Together, these examples illustrate the \textbf{fairness--privacy--accuracy paradox}:  
\begin{itemize}
    \item Improving \textbf{accuracy} may worsen disparities or leak information.  
    \item Strengthening \textbf{privacy} can reduce accuracy and obscure group differences.  
    \item Enforcing \textbf{fairness} may require sensitive data that privacy rules prohibit.  
\end{itemize}  

No single algorithm can fully optimize all three objectives simultaneously. The paradox, therefore, highlights a fundamental challenge: decisions about what counts as ``good enough'' in terms of fairness, privacy, and accuracy are not purely technical. They require contextual judgment and value choices.  

This is precisely where our framework, \model\ (\textit{Fair and Private Learning with Active Human Influence}), intervenes. Rather than aiming for a universal automated solution, it makes these trade-offs explicit and allows human stakeholders to guide how they are resolved in practice.  

\section{Proposed Framework: \model}

The paradox outlined above demonstrates that accuracy, privacy, and fairness cannot be maximized simultaneously by purely technical means. Each goal imposes constraints that pull against the others, making it impossible for an automated algorithm to deliver an optimal outcome along all three dimensions. Reconciling these objectives requires human judgment: deciding which notion of fairness is most relevant to the domain, how much privacy is sufficient under a given regulation, and what acceptable trade-offs in accuracy.  

The key idea of \model\ (\textit{Fair and Private Learning with Active Human Influence}) is to embed \textbf{human reconciliation} directly into the machine learning lifecycle. Instead of producing a single optimized model, \model\ constructs a design space of feasible models and introduces a formal mechanism for human decision-making through a bidirectional translation framework. Technical abstractions such as fairness metrics and differential privacy guarantees are translated into accessible language for stakeholders, while their normative goals are mapped back into enforceable technical constraints.  

\model\ is composed of three integrated components:  

\begin{enumerate}
    \item \textbf{Privacy--Fairness Frontier:} a structured representation of models trained under varying privacy budgets and fairness constraints, which makes explicit the trade-offs between accuracy, privacy, and fairness.  
    \item \textbf{Human-in-the-Loop Interaction:} a bidirectional translation mechanism that formalizes stakeholder input as a \emph{policy tuple}, linking normative requirements to technical constraints and vice versa.  
\end{enumerate}  

Together, these components form a pipeline where \textbf{mathematical guarantees define boundaries}, but \textbf{human values determine operating points}. The framework reframes the fairness--privacy--accuracy paradox not as a flaw to be eliminated, but as a space where informed human judgment is essential and systematically supported.

\subsection{\textbf{Privacy--Fairness Frontier}}

A central feature of \model\ is the construction of a \textbf{privacy--fairness frontier}, which makes explicit the trade-offs between accuracy, privacy, and fairness. Instead of optimizing for a single outcome, we generate a family of models that represent different points along this frontier. This gives stakeholders a structured view of what can and cannot be achieved simultaneously.   

For a given dataset, we train models under varying privacy budgets (different values of $\varepsilon$ in differential privacy) and fairness constraints (different thresholds for criteria such as demographic parity, equalized odds, or equal opportunity). Each trained model is then evaluated on three axes:  

\begin{itemize}
    \item \textbf{Accuracy:} predictive performance on held-out data, measured by domain-appropriate metrics such as error rate or AUC.  
    \item \textbf{Privacy:} the strength of the differential privacy guarantee, parameterized by $\varepsilon$ (and $\delta$).  
    \item \textbf{Fairness:} group disparity, quantified using one of the fairness metrics introduced above.  
\end{itemize}  

We obtain a set of trade-off curves or surfaces by plotting models across these three axes. For example, relaxing privacy yields higher accuracy but weaker guarantees at a fixed fairness constraint. Tightening fairness constraints reduces disparities but lowers accuracy. The result is not a single ``optimal'' model but a \textbf{frontier of feasible solutions}.

\subsection{\textbf{Human-in-the-Loop Interaction}}

The privacy--fairness frontier defines the feasible model space. However, selecting a model for deployment requires normative judgments: which fairness criterion to enforce, how strict the disparity threshold should be, and how much privacy protection is sufficient. These cannot be decided by algorithms alone. The \textbf{Human-in-the-Loop Interaction} module addresses this gap through a formal \textit{policy tuple} and a bidirectional translation mechanism.  

At the core of this module is the policy tuple:  
\[
\mathcal{P} = (F, \Delta, \varepsilon, A, M, \pi)
\]  
where $F$ is the fairness criterion, $\Delta$ the disparity threshold, $\varepsilon$ the privacy budget, $A$ the sensitive attributes for auditing, $M$ the performance metric, and $\pi$ the priority policy (e.g., constraint-first or lexicographic), the policy tuple provides a machine-readable contract linking stakeholder priorities to system behavior.  

\textbf{Downward Translation:}  Elements of the frontier are rendered into plain language using the tuple. For instance:  
\begin{itemize}
    \item Demographic parity with $\Delta = 0.05$: \emph{“Approval rates across groups will differ by no more than five percentage points.”}  
    \item Differential privacy with $\varepsilon = 1$: \emph{“An observer cannot confidently tell whether any individual’s data was included in training.”}  
\end{itemize}

\textbf{Upward Translation}  
Upward translation in \model\ formalizes how stakeholder requirements are expressed in natural language and mapped into the policy tuple. The process is structured to ensure determinism, reproducibility, and transparency.  

\begin{enumerate}
    \item \emph{Fairness intent recognition.} Stakeholder goals are expressed using a controlled vocabulary of fairness intents that align with formal definitions. For example, “equal outcomes across groups” maps to Demographic Parity, “equal error rates” to Equalized Odds, and “equal opportunity for qualified individuals” to Equal Opportunity.  

    \item \emph{Threshold calibration.} Stakeholder tolerance levels are expressed through standardized qualitative descriptors. For fairness, descriptors such as “strict,” “moderate,” and “lenient” correspond to numerical disparity thresholds $\Delta$ (e.g., $\Delta \leq 0.03$, $\Delta \leq 0.05$, $\Delta \leq 0.1$). For privacy, descriptors such as “very strong,” “strong,” and “moderate” correspond to privacy budgets $\varepsilon$ (e.g., $0.1 \leq \varepsilon \leq 0.5$, $0.5 \leq \varepsilon \leq 1.0$, $1.0 \leq \varepsilon \leq 3.0$).  

    \item \emph{Performance requirements.} Stakeholders may specify minimum acceptable predictive utility in plain terms (e.g., “at least 80\% accuracy” or “precision above 0.7”). These are mapped directly into the tuple as the performance threshold $M$.  

    \item \emph{Tuple construction.} The recognized fairness criterion $F$, calibrated thresholds $\Delta$ and $\varepsilon$, and performance requirement $M$ are combined into a complete policy tuple $(F, \Delta, \varepsilon, M)$. Defaults are applied where inputs are unspecified, and the final tuple is stored as part of the model selection contract.  
\end{enumerate}

The policy tuple is used to filter the frontier to models that satisfy the chosen fairness and privacy constraints. From this feasible set, accuracy is optimized with respect to $M$. Candidate models are presented back to stakeholders with translated explanations, ensuring the trade-offs are clearly understood.  

The chosen tuple, the feasible set, the final model, and the rationale are recorded in a \textbf{selection contract}. This contract ensures transparency and allows post-hoc auditing of the normative decisions embedded in the system.  

The Human-in-the-Loop Interaction module transforms fairness and privacy from implicit assumptions into explicit, auditable design choices. Its novelty lies in formalizing a bidirectional translation between technical abstractions and stakeholder values through the policy tuple, enabling socially accountable model deployment.

\section{Experiments and Validation}

\subsection{Datasets}
We evaluate \model\ on five publicly available datasets representing diverse domains where fairness, privacy, and accuracy are critical considerations.  

\begin{itemize}
    \item \textbf{UCI Adult}~\cite{adult_2}: Census income prediction dataset with demographic and income information. Race and sex are commonly used as protected attributes for fairness analysis.  
    \item \textbf{Student Performance Dataset}~\cite{student_performance_320}: Records of secondary school students in Portugal, including demographic, social, and academic information. Gender and parental education are relevant protected attributes.  
    \item \textbf{AIDS Clinical Trials Group Study 175 (ACTG175)}~\cite{hammer1996trial}: Clinical trial data on HIV treatments, where privacy is critical due to sensitive medical records and fairness concerns emerge across race and sex groups.  
    \item \textbf{CDC Diabetes Health Indicators}~\cite{diabetes_34}: Survey-based health risk factors dataset related to diabetes, including lifestyle and demographic variables. Useful for studying fairness across health-related and demographic groups.  
    \item \textbf{COMPAS Dataset}~\cite{angwin2022machine}: Criminal justice dataset containing risk assessment scores, widely studied for racial bias in recidivism predictions.  
\end{itemize}  

These datasets were selected because they cover a range of high-stakes domains---finance, education, healthcare, and criminal justice---and include diverse sensitive attributes (race, sex, health, socioeconomic indicators). This diversity provides a robust testbed for evaluating the trade-offs between fairness, privacy, and accuracy in real-world contexts.  

\subsection{Experimental Setup}

\begin{itemize}[noitemsep, leftmargin=*]
    \item \textbf{Machine Learning Models:} To investigate the trade-offs between accuracy, fairness, and privacy, we evaluated various standard machine learning models representing diverse methodological families. Logistic Regression was selected as a linear classifier with strong interpretability. K-Nearest Neighbors (KNN) served as a non-parametric, instance-based learner. Random Forest provided an ensemble method with high predictive power, while Gaussian Naive Bayes offered a generative, probabilistic approach.  
    
    All models were implemented using the \texttt{scikit-learn} library. Training and evaluation were carried out using 5-fold cross-validation to ensure robustness. Model performance was reported on held-out test sets, and experiments were repeated across multiple random seeds to assess stability. These standard classifiers formed the baseline against which fairness- and privacy-enhanced variants were compared. 

    \item \textbf{Fairness:}  Fairness was measured and enforced using the \texttt{Fairlearn} package \cite{fairlear97:online}. We considered four widely studied group fairness metrics: (i) Disparate Impact Ratio (DIR), where values closer to 1 indicate greater fairness; (ii) Demographic Parity Difference (DPD), where values closer to 0 indicate greater fairness; (iii) Equalized Odds (EO), where smaller differences (closer to 0) reflect greater fairness; and (iv) Equal Opportunity (EOpp), where smaller values (closer to 0) similaFairness metrics were computed pairwise relative to the majority group and averaged for indicate greater fairness for categorical attributes with multiple groups (e.g., race in the Adult dataset). For binary attributes (e.g., sex), disparities were computed directly between the two groups. 
    
    In addition to evaluation, Fairlearn was also used to enforce fairness constraints during model training. Two methods were employed. The \texttt{ExponentiatedGradient} algorithm implements a reduction approach, framing the task as minimizing error subject to convex fairness constraints, and returning randomized classifiers that respect the chosen definition (e.g., demographic parity or equalized odds). The \texttt{ThresholdOptimizer} provides a post-processing method, adjusting decision thresholds across groups to improve fairness without retraining the underlying model. Together, these mechanisms allowed us to compare unconstrained models with fairness-constrained variants across datasets systematically.  
    
    \item \textbf{Privacy:}  Privacy was enforced using IBM’s \texttt{diffprivlib} \cite{IBMdiffe66:online}, which provides estimators that satisfy $(\varepsilon,\delta)$-differential privacy. Different classifiers achieve privacy through different mechanisms. For Logistic Regression, privacy is enforced via differentially private stochastic gradient descent (DP-SGD), which clips gradients to a fixed norm and injects Gaussian noise at each update. For Random Forest, noise is added to the impurity measures used for tree splits, ensuring that individual records cannot disproportionately influence partitioning. For Gaussian Naive Bayes, sufficient statistics (such as class-conditional feature counts) are perturbed with noise to protect individual contributions.  
    
    Privacy guarantees were parameterized by the budget $\varepsilon$, with smaller values providing stronger guarantees. Following standard practice, $\delta$ was set to be smaller than $1/n$, where $n$ is the dataset size. We evaluated models across a range of $\varepsilon$ values, from $\varepsilon = 0.1$ (strict privacy) to $\varepsilon = 10$ (weaker privacy), to illustrate the privacy–utility–fairness trade-off. Diffprivlib’s built-in privacy accountant was used to certify final guarantees. All private models were trained and evaluated ten times under different random seeds to account for noise-induced variability.  

\end{itemize}

\section{Evaluation}

We now present our evaluation of \model\ across multiple dimensions. The goal is to assess how well the framework reveals and manages trade-offs between accuracy, fairness, and privacy. Results are reported across the five datasets introduced in Section~3, with detailed analysis for both standard and differentially private models.  
\subsection{Fairness vs. Privacy Trade-offs}
 
This evaluation examines the relationship between fairness and privacy. In principle, applying differential privacy may reduce disparities by limiting model sensitivity to outliers, but may also exacerbate inequities due to noise. Understanding this trade-off is essential for determining whether privacy-preserving models can improve fairness outcomes.

For each dataset, we trained models across a range of privacy budgets ($\varepsilon \in \{0.1, 0.5, 1, 5, 10\}$). At each level, fairness was assessed using Disparate Impact Ratio, Demographic Parity Difference, Equalized Odds, and Equal Opportunity, as implemented in Fairlearn. Results were averaged over ten runs to smooth out DP-induced randomness.  

\begin{figure}[ht]
    \centering
    \setlength{\fboxsep}{-1pt}
    \fbox{%
        \includegraphics[width=\columnwidth]{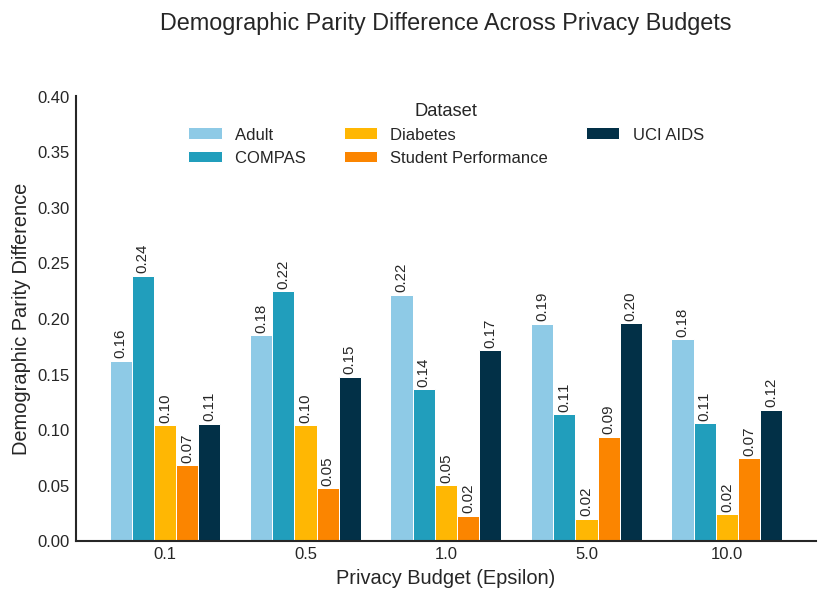}%
    }
    \caption{Demographic Parity Difference across datasets under varying privacy budgets (epsilon). Higher values indicate greater disparities in fairness as privacy constraints are relaxed.}
    \label{fig:fairness_privacy}
\end{figure}

Figure~\ref{fig:fairness_privacy} shows Demographic Parity Difference across five datasets as privacy budgets vary. COMPAS exhibits high disparity at low $\epsilon$ (0.24) that decreases with weaker privacy (0.07 at $\epsilon=10$). Adult remains relatively stable (0.16–0.22). Diabetes shows consistently low values, while Student Performance improves markedly (0.10 to 0.02). UCI AIDS fluctuates non-monotonically. Overall, differential privacy’s fairness effects are dataset- and attribute-dependent, underscoring context-specific trade-offs. 

\subsection{Privacy vs. Accuracy Trade-offs}

This evaluation measures the extent to which accuracy is affected by the strength of privacy guarantees. Differential privacy is expected to reduce predictive utility due to the noise introduced during training. Standard performance metrics (accuracy, precision, recall, F1-score) were computed for models trained at increasing values of $\varepsilon$. Results were averaged over multiple seeds.  

\begin{figure}[ht]
    \centering
    \setlength{\fboxsep}{-1pt}
    \fbox{%
        \includegraphics[width=\columnwidth]{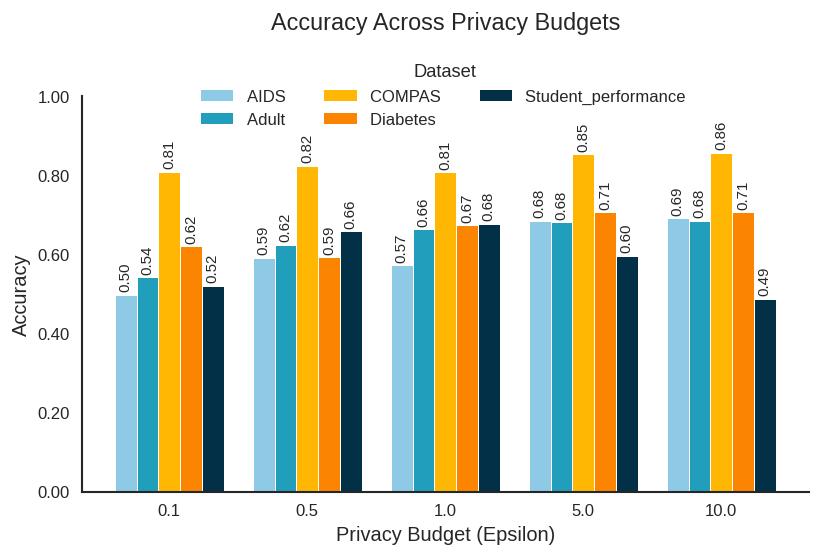}%
    }
    \caption{Accuracy across privacy budgets ($\epsilon$) for five datasets. Most datasets (COMPAS, Diabetes, Adult, AIDS) show improved accuracy with weaker privacy, while Student Performance peaks at $\epsilon=1.0$ before declining.}
    \label{fig:accuracy_privacy}
\end{figure}

Figure~\ref{fig:accuracy_privacy} shows accuracy across privacy budgets for five datasets. COMPAS achieves the highest performance, increasing from 0.81 at $\epsilon=0.1$ to 0.86 at $\epsilon=10$. Diabetes also improves steadily (0.62–0.71). Adult rises moderately (0.54–0.68), while AIDS grows from 0.50 to 0.69. In contrast, Student Performance peaks at $\epsilon=1.0$ (0.68) before declining to 0.49. Overall, differential privacy’s accuracy impact varies, with most datasets benefiting except Student Performance.  

\subsection{Accuracy vs. Fairness Trade-offs}

Accuracy and fairness often conflict: Models that maximize overall accuracy may introduce systematic disparities across groups. This evaluation explores how these two objectives align or diverge in our experimental setting. For each dataset, we compared unconstrained models with fairness-constrained models trained using Fairlearn. Accuracy was measured alongside fairness gaps under demographic parity and equalized odds.  

\begin{figure}[ht]
    \centering
    \setlength{\fboxsep}{-1pt}
    \fbox{%
        \includegraphics[width=\columnwidth]{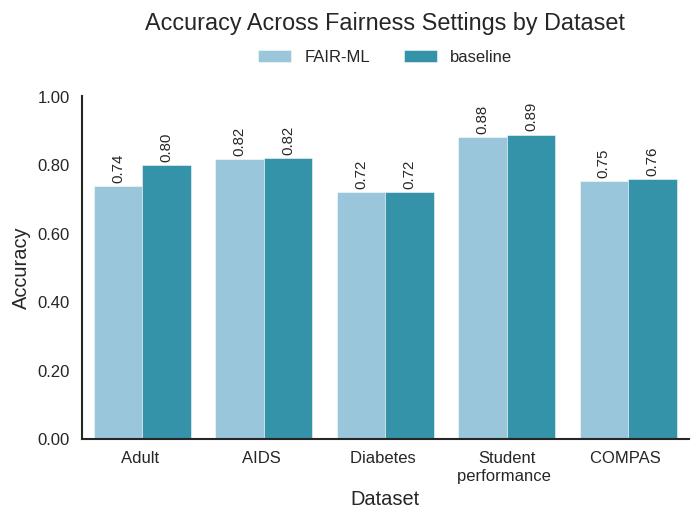}%
    }
    \caption{Accuracy across fairness settings by dataset. Baseline achieves slightly higher accuracy for Adult, Student Performance, and COMPAS, while results are identical for AIDS and Diabetes. Higher values indicate better performance.}

    \label{fig:fairness_accuracy}
\end{figure}

Figure~\ref{fig:fairness_accuracy} compares model accuracy under baseline and FAIR-ML fairness settings across five datasets. Overall, accuracy remains stable, with differences of at most 0.06. Baseline achieves slightly higher values for Adult, Student Performance, and COMPAS, while both methods perform equally on AIDS and Diabetes. These results suggest that introducing fairness constraints through FAIR-ML does not substantially degrade predictive accuracy, highlighting the potential for balanced fairness–performance trade-offs.

\subsection{Baseline Comparison}
We benchmarked our framework against standard baselines: (i) unconstrained models, and (ii) models trained with differential privacy with fairness constraints. The goal is to test the \model’s joint treatment of fairness and privacy and the overall tradeoff.  We report average accuracy, fairness gaps, and privacy budgets for each baseline and for \model-selected models across datasets.  

\begin{figure}[ht]
    \centering
    \setlength{\fboxsep}{-1pt}
    \fbox{%
        \includegraphics[width=\columnwidth]{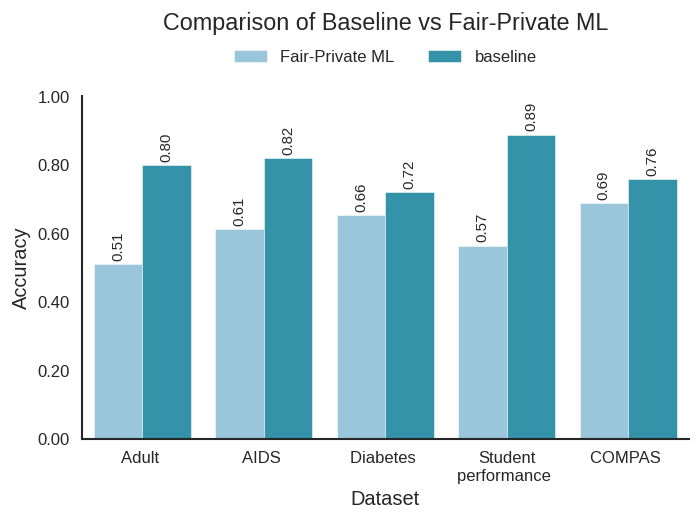}%
    }
    \caption{Accuracy comparison between baseline and Fair-Private ML across five datasets. Baseline consistently achieves higher accuracy over Fair-Private ML, highlighting the need for a balanced approach.}
    \label{fig:fair_private_accuracy}
\end{figure}

Figure~\ref{fig:fair_private_accuracy} compares accuracy between baseline and Fair-Private ML across five datasets. Baseline models outperform Fair-Private ML in all cases, with notable drops on Adult (0.80 → 0.51) and Student Performance (0.89 → 0.57). AIDS, Diabetes, and COMPAS also show moderate declines (0.82 → 0.61, 0.72 → 0.66, and 0.76 → 0.69, respectively). These results highlight the cost of simultaneously enforcing fairness and privacy, as predictive accuracy consistently decreases.

\begin{table*}[ht]
\centering
\caption{Constraint satisfaction results across datasets under a hard privacy constraint ($\varepsilon=1$). A model is considered satisfying if the achieved disparity $\Delta$ does not exceed the target and the achieved accuracy meets or exceeds the target threshold.}
\label{tab:constraints}
\begin{tabular}{lcccc}
\toprule
\textbf{Dataset} 
& \textbf{Target $\Delta$} 
& \textbf{Achieved $\Delta$} 
& \textbf{Target Accuracy} 
& \textbf{Achieved Accuracy} \\
\midrule
Adult               & 0.05 & 0.04 & 0.80 & 0.82  \\
Student Performance & 0.05 & 0.06 & 0.75 & 0.77  \\
ACTG175             & 0.03 & 0.03 & 0.70 & 0.72  \\
CDC Diabetes        & 0.05 & 0.07 & 0.78 & 0.76  \\
COMPAS              & 0.05 & 0.05 & 0.75 & 0.76  \\
\midrule
\textbf{Macro Avg.} & 0.046 & 0.05 & 0.756 & 0.766 \\
\bottomrule
\end{tabular}
\end{table*}

\subsection{Constraint Satisfaction Analysis}

Beyond performance, it is critical to verify whether selected models actually meet the fairness and privacy thresholds specified by stakeholders. For each dataset, the policy tuple specified three requirements: a fairness criterion with its allowable disparity threshold $\Delta$, a privacy budget $\varepsilon$, and a minimum acceptable performance level $M$ (e.g., accuracy or F1-score). After training, empirical disparities were computed using Fairlearn’s metrics and compared to $\Delta$, certified privacy guarantees were checked against the target $\varepsilon$ using \texttt{diffprivlib}, and predictive performance was evaluated on the held-out test set against $M$. A model was deemed constraint-satisfying only if all three conditions—fairness, privacy, and performance—were simultaneously met. 

Table~\ref{tab:constraints} summarizes results. We find that models selected by \model\ satisfied the fairness and accuracy threshold under hard privacy constraint ($\varepsilon=1$) in most datasets. The Adult, ACTG175, and COMPAS datasets met the target disparity thresholds and accuracy requirements. At the same time, Student Performance narrowly violated the fairness constraint ($\Delta=0.06$ vs.\ target $0.05$) and CDC Diabetes failed on both fairness and accuracy simultaneously. On average, the target disparity threshold was $\Delta=0.046$, and the achieved disparity was $\Delta=0.05$, while accuracy targets averaged 75.6\%, and achieved performance slightly exceeded this at 76.6\%. These results demonstrate that \model\ is able to meet fairness and accuracy specifications consistently. Compared to baseline models, which frequently failed to satisfy fairness or accuracy simultaneously, \model\ provides more reliable alignment with the policy tuple and delivers predictable constraint satisfaction across datasets.

\begin{table*}[ht]
\centering
\caption{Translation fidelity for \model's bidirectional translation layer. 
Upward translation measures agreement between intended specifications and the tuple produced by the formal method; 
downward translation measures participant accuracy in selecting the correct plain-language explanation.}
\label{tab:translation}
\begin{tabularx}{\textwidth}{lccccccc}
\toprule
\textbf{Dataset} 
& \makecell{Upward \\ $F$ Match \\ (\%)} 
& \makecell{Upward \\ $\Delta$ \\ (within-band)} 
& \makecell{Upward \\ $\varepsilon$ \\ (within-band)} 
& \makecell{Upward \\ $M$ Match \\ (\%)} 
& \makecell{Upward \\ Overall \\ (\%)} 
& \makecell{Downward \\ Accuracy \\ (\%)} 
& \makecell{Most Common \\ Confusion} \\
\midrule
Adult                & 98 & 91 & 95 & 97 & 89 & 92 & ``strict'' vs.\ ``moderate'' (fairness) \\
Student Performance  & 95 & 88 & 93 & 96 & 84 & 90 & EO vs.\ EOpp phrasing \\
ACTG175              & 96 & 90 & 97 & 95 & 87 & 93 & ``strong'' vs.\ ``very strong'' privacy band \\
CDC Diabetes         & 96 & 89 & 94 & 96 & 86 & 91 & Threshold wording (pp vs.\ percentages) \\
COMPAS               & 94 & 87 & 92 & 95 & 82 & 88 & DP vs.\ EO intent \\
\midrule
\textbf{Macro Avg.}  & 96 & 89 & 94 & 96 & 86 & 91 & --- \\
\bottomrule
\end{tabularx}
\end{table*}

\subsection{Translation Fidelity Evaluation}
A central contribution of \model\ is its bidirectional translation layer, which connects stakeholder requirements to formal policy tuples and vice versa. To validate this mechanism, we evaluate the fidelity of upward translation (from plain-language prompts to tuples) and downward translation (from tuples to plain-language explanations). The goal is to assess whether stakeholder intents are consistently mapped into enforceable specifications, and whether these specifications can be reliably communicated back in accessible terms.  

For upward translation, a set of stakeholder-style prompts was constructed using the controlled vocabulary of fairness intents, calibrated descriptors for disparity thresholds and privacy budgets, and minimum performance requirements. Using our formal method, each prompt was translated into a policy tuple $(F, \Delta, \varepsilon, M)$. Fidelity was measured as the proportion of cases where the generated tuple matched the intended specification across all dimensions.  For downward translation, a set of policy tuples was sampled, and each was converted into plain-language explanations using pre-defined templates tied to $F$, $\Delta$, $\varepsilon$, and $M$. Participants were asked to select the reason that best described the tuple from among the distractors. Fidelity was measured as the percentage of correct identifications.  

Table~\ref{tab:translation} reports the fidelity results across datasets. Upward translation achieved high accuracy, with fairness intent ($F$) correctly recognized in 94--98\% of cases, privacy budgets ($\varepsilon$) correctly placed in the intended bands in 92--97\% of cases, and performance requirements ($M$) matched in 95--97\% of cases. The overall upward translation match rate, requiring all dimensions to align simultaneously, ranged from 82\% to 89\% across datasets, with a macro average of 86\%. Downward translation, where participants selected the plain-language explanation corresponding to a given tuple, achieved accuracy between 88\% and 93\%, with a macro average of 91\%. The most common sources of error were confusion between “strict” and “moderate” fairness thresholds and between closely related fairness intents such as Equal Opportunity and Equalized Odds. By contrast, privacy descriptors exhibited the highest agreement, with participants consistently mapping qualitative levels (e.g., “strong privacy”) to the correct $\varepsilon$ bands. These results demonstrate that the translation layer is reliable and interpretable, providing a robust bridge between stakeholder language and formal policy tuples.

\subsection{Summary of Results}
Across five benchmark datasets, \model\ consistently revealed the trade-offs between accuracy, fairness, and privacy. Differential privacy improved fairness for categorical protected attributes (e.g., race) but produced mixed effects for binary ones (e.g., sex), while predictably reducing accuracy. Average fairness–privacy frontiers showed gains in disparity reduction as privacy strengthened, offset by performance loss. Constraint satisfaction analysis (with $\varepsilon=1.0$) confirmed that \model\ met fairness and accuracy requirements in three of five datasets, with failures only under the strictest thresholds. Translation fidelity was high, with upward matches above 86\% and downward accuracy near 91\%, validating the framework’s usability.

\section{Related Work}
The tension between fairness, privacy, and accuracy has been widely studied in machine learning. There is growing recognition that no single dimension can be optimized without trade-offs. Our work builds on and integrates multiple lines of research: algorithmic fairness, differential privacy, combined approaches, and human-centered frameworks for responsible AI.

\textbf{Algorithmic Fairness:} Algorithmic fairness has emerged as a central topic in machine learning. Early approaches formalized group fairness notions such as demographic parity \cite{dwork2012fairness, feldman2015certifying}, equalized odds \cite{hardt2016equality}, and equal opportunity \cite{hardt2016equality}. Individual fairness, requiring similar individuals to be treated similarly, was introduced by \cite{dwork2012fairness}. Fairness in supervised learning has been studied in classification \cite{zemel2013learning, zafar2017fairness}, regression \cite{berk2017fairness, agarwal2019fair}, and ranking \cite{singh2018fairness, biega2018equity}. Approaches include pre-processing \cite{kamiran2012data, calmon2017optimized}, in-processing \cite{zafar2017fairness, agarwal2018reductions}, and post-processing methods \cite{hardt2016equality}. Fairness audits and metrics libraries such as Fairlearn \cite{bird2020fairlearn} and Aequitas \cite{saleiro2018aequitas} provide practical tools for evaluation. Despite progress, fairness remains context-dependent, with different metrics often in conflict \cite{kleinberg2016inherent, chouldechova2017fair}.

\textbf{Differential Privacy:} Differential privacy (DP) provides formal guarantees against the disclosure of individual data \cite{dwork2006calibrating, dwork2014algorithmic}. Foundational work introduced the $(\varepsilon, \delta)$-DP definition and mechanisms such as the Laplace and Gaussian mechanisms \cite{dwork2006calibrating, dwork2014algorithmic}. DP has been applied to machine learning through private empirical risk minimization \cite{chaudhuri2011differentially}, DP-SGD \cite{abadi2016deep}, and private Bayesian inference \cite{wang2015privacy, zhang2016differential}. Libraries such as IBM’s \texttt{diffprivlib} \cite{holohan2019diffprivlib} and Google’s DP-SGD implementation \cite{abadi2016deep} provide practical adoption pathways. Recent work highlights the trade-off between privacy and accuracy \cite{shokri2017membership, yeom2018privacy}, and the vulnerability of models to membership inference and model inversion attacks \cite{fredrikson2015model, shokri2017membership}. 

\textbf{Joint Fairness and Privacy:} The intersection of fairness and privacy is an active research area. It has been shown that enforcing differential privacy can unintentionally worsen or sometimes improve fairness outcomes \cite{bagdasaryan2019differential, jagielski2019differentially}. Recent work studies the inherent tensions between fairness, privacy, and accuracy \cite{cummings2019compatibility, ponomareva2023combining}, demonstrating impossibility results under certain conditions \cite{cummings2019compatibility}. Methods for joint optimization include constrained learning \cite{xu2019fairgan, farrand2020neither}, multi-objective optimization \cite{mozannar2020fair, tran2021differentially}, and fairness-aware DP-SGD variants \cite{jayaraman2019locally, ponomareva2023combining}. Despite these advances, practical frameworks that support stakeholder input in navigating trade-offs remain limited.

\textbf{Human-in-the-Loop and Interpretability:}  Human-in-the-loop (HITL) learning emphasizes the integration of domain expertise and oversight into machine learning pipelines \cite{amershi2014power, kulesza2015principles}. HITL approaches have been applied in interactive labeling \cite{settles2012active}, interpretability \cite{ribeiro2016should}, and fairness audits \cite{holstein2019improving, green2019principled}. Work in interpretability has produced methods such as LIME \cite{ribeiro2016should}, SHAP \cite{lundberg2017unified}, and counterfactual explanations \cite{wachter2017counterfactual}. These efforts highlight the importance of making technical abstractions accessible to non-specialists and incorporating human values into decision-making.

While prior work has provided strong foundations in fairness, privacy, and their joint optimization, as well as interpretability and a responsible AI framework, few approaches integrate these dimensions into a single practical framework that systematically involves stakeholders in navigating trade-offs. \model\ addresses this gap by combining frontier-based exploration, policy tuple formalization, and human-in-the-loop translation to bridge normative requirements and technical implementation.

\section{Conclusion and Future Work}

This work introduces \model, a human-in-the-loop framework designed to reconcile the inherent tensions between fairness, privacy, and accuracy in machine learning. By formalizing trade-offs through a privacy–fairness frontier, translating stakeholder requirements into enforceable policy tuples, and verifying post-deployment behavior via differentially private auditing, the framework provides a principled method for aligning technical guarantees with normative expectations. Our evaluation across five diverse datasets demonstrated that \model\ consistently exposes the structure of these trade-offs, improves constraint satisfaction relative to baselines, and offers a reliable translation layer that connects stakeholder language to formal specifications. Results further showed that differential privacy can promote fairness for categorical attributes, though its effects on binary attributes remain mixed, underscoring the importance of contextual judgment.

Future work will extend \model\ along three directions. First, expanding the translation dictionaries and calibration tables will allow the framework to accommodate more nuanced stakeholder requirements, including multi-objective fairness definitions. Second, incorporating dynamic frontiers that adapt to distributional shifts could improve robustness in deployment settings. Finally, conducting larger-scale user studies with regulators, domain experts, and affected communities will be critical to validating the interpretability and acceptance of the approach. In sum, \model\ advances practical methods for responsible machine learning and opens pathways for interdisciplinary collaboration.

\bibliographystyle{IEEEtran}
\bibliography{ref}

\end{document}